\documentclass{article}

\usepackage{microtype}
\usepackage{graphicx}
\usepackage{subfigure}
\usepackage{booktabs} 

\usepackage{hyperref}

\usepackage[accepted]{icml2025}

\newcommand{\argmax}{\mathop{\mathrm{argmax}}}
\newcommand{\argmin}{\mathop{\mathrm{argmin}}}
\newcommand{\E}{\mathbb{E}}

\newcommand{\Pcal}{\mathcal P}

\newcommand{\Xcal}{\mathcal X}
\newcommand{\Ycal}{\mathcal Y}

\newcommand{\Ocal}{\mathcal O}

\newcommand{\Scal}{\mathcal{S}}

\newcommand{\inner}[1]{\left\langle#1\right\rangle}

\renewcommand{\P}{\mathbb{P}}

\newcommand{\bra}[1]{\left[#1\right]}
\newcommand{\pa}[1]{\left(#1\right)}

\newcommand{\KL}{\mathrm{KL}}

\newcommand{\wh}{\widehat}

\usepackage{amsmath}
\usepackage{amssymb}
\usepackage{mathtools}
\usepackage{amsthm}

\usepackage[capitalize,noabbrev]{cleveref}

\theoremstyle{plain}
\newtheorem{theorem}{Theorem}[section]

\theoremstyle{definition}
\newtheorem{definition}[theorem]{Definition}

\theoremstyle{remark}

\usepackage[textsize=tiny]{todonotes}

\icmltitlerunning{Improving LLM General Preference Alignment via Optimistic Online Mirror Descent}

\begin{document}

\twocolumn[
\icmltitle{Improving LLM General Preference Alignment \\ via Optimistic Online Mirror Descent}

\icmlsetsymbol{equal}{*}

\begin{icmlauthorlist}
\icmlauthor{Yuheng Zhang}{uiuc}
\icmlauthor{Dian Yu}{tencent}
\icmlauthor{Tao Ge}{tencent}
\icmlauthor{Linfeng Song}{tencent}
\icmlauthor{Zhichen Zeng}{uiuc}
\icmlauthor{Haitao Mi}{tencent}
\icmlauthor{Nan Jiang}{uiuc}
\icmlauthor{Dong Yu}{tencent}
\end{icmlauthorlist}

\icmlaffiliation{uiuc}{University of Illinois Urbana-Champaign}
\icmlaffiliation{tencent}{
Tencent AI Lab, Bellevue}

\icmlcorrespondingauthor{Yuheng Zhang}{yuhengz2@illinois.edu}

\icmlkeywords{LLM Alignment, RLHF}

\vskip 0.3in
]



\printAffiliationsAndNotice{}  

\begin{abstract}
Reinforcement learning from human feedback (RLHF) has demonstrated remarkable effectiveness in aligning large language models (LLMs) with human preferences. Many existing alignment approaches rely on the Bradley-Terry (BT) model assumption, which assumes the existence of a ground-truth reward for each prompt-response pair. However, this assumption can be overly restrictive when modeling complex human preferences. In this paper, we drop the BT model assumption and study LLM alignment under general preferences, formulated as a two-player game. Drawing on theoretical insights from learning in games, we integrate optimistic online mirror descent into our alignment framework to approximate the Nash policy. Theoretically, we demonstrate that our approach achieves an $\mathcal{O}(T^{-1})$ bound on the duality gap, improving upon the previous $\mathcal{O}(T^{-1/2})$ result. More importantly, we implement our method and show through experiments that it outperforms state-of-the-art RLHF algorithms across multiple representative benchmarks.
\end{abstract}

\section{Introduction}\label{sec:intro}
Reinforcement learning from human feedback (RLHF) has played a pivotal role in aligning large language models (LLMs) with human preferences. The goal of RLHF is to fine-tune LLMs to generate responses that are preferred by humans. It has been successfully deployed in state-of-the-art models, including Instruct-GPT~\citep{ouyang2022training} and Claude~\citep{bai2022training}. The first RLHF framework for LLMs was developed by~\citet{ouyang2022training}, where after the pre-training stage, the LLM is fine-tuned to maximize the reward signal from a reward model using the proximal policy optimization (PPO) algorithm~\citep{schulman2017proximal}. This pipeline requires training both the reward model and the policy model. In addition, policy gradient approaches such as PPO often exhibit high variance and instability during training~\citep{peng2023stabilizing}, leading to increased computational costs.

To develop a more stable and computationally lightweight alignment approach,~\citet{rafailov2024direct} propose the Direct Preference Optimization (DPO) algorithm, which directly trains the LLM on a preference dataset and bypasses the need for a reward model. DPO uses an offline preference dataset, and since its development, a line of research has explored different exploration strategies and proposed online direct preference alignment algorithms~\citep{xiong2024iterative,xie2024exploratory,dong2024rlhf,yuan2024self}. All these methods assume that human preferences can be modeled using the Bradley-Terry (BT) model, where a reward function $R^*$ exists such that, for any prompt $x$ and response pair $(y^1, y^2)$, the preference between $y^1$ and $y^2$ satisfies:  
\[
\mathbb{P}(y^1 \succ y^2 \mid x) = \sigma(R^*(x,y^1) - R^*(x,y^2)),
\]
where $\sigma(z) = \frac{1}{1+\exp(-z)}$ is the sigmoid function.

However, the existence of a reward function and the BT model are strong assumptions that can be overly restrictive when modeling complex human preferences. For example, the preference signals in the BT model are always transitive: if $A$ is preferred to $B$ and $B$ is preferred to $C$, then $A$ must always be preferred to $C$. This transitive property contradicts evidence from human decision-making~\citep{may1954intransitivity,tversky1969intransitivity}, especially when preferences are at the population level and aggregated from different human groups~\citep{may1954intransitivity,ye2024theoretical}.  Furthermore, the limitations of the BT model have also been observed in RLHF practice. ~\citet{jiang2023llm} show that a preference model with 0.4B parameters achieves performance comparable to Llama-2-13B-based reward models. ~\citet{ye2024theoretical} train a BT reward model and a preference model separately using the same base model and preference dataset, and their results demonstrate that the preference model consistently outperforms the reward model on Reward-Bench~\citep{lambert2024rewardbench} under both base models. These findings motivate us to drop the BT model assumption and instead consider general preferences.

In this work, we study the problem of aligning LLMs with general preferences and formulate it as a two-player zero-sum game. Our objective is to approximate the Nash policy of the game, which ensures a win rate of at least 50\% against any other policy. As established in the game theory literature~\citep{bai2020near,liu2021sharp}, self-play algorithms have proven to be highly effective in approximating Nash policies. Building on this, we aim to propose a novel online RLHF algorithm that further leverages the self-play strcture to enhance general preference alignment for LLMs. Our contributions are summarized as follows.

\paragraph{Contributions.} We propose a novel online general preference alignment algorithm, Optimistic Nash Policy Optimization (ONPO). Inspired by recent advancements in game theory, our algorithm integrates optimistic online mirror descent~\citep{rakhlin2013optimization,syrgkanis2015fast} into the self-play framework. By utilizing a reward predictor in a two-step update strategy, ONPO more effectively leverages the self-play mechanism and achieves a faster convergence rate of $\mathcal{O}(T^{-1})$, improving upon the previous $\mathcal{O}(T^{-1/2})$ result.  

ONPO can be efficiently implemented by directly minimizing a loss objective on a preference dataset, making it computationally lightweight in practice. We evaluate ONPO on several representative benchmarks, comparing it with state-of-the-art general preference alignment algorithms. Experimental results demonstrate that ONPO consistently outperforms or achieves performance comparable to the baselines across different base models and benchmarks. Notably, on the AlpacaEval 2.0 benchmark~\citep{li2023alpacaeval}, ONPO achieves a 21.2\% and 9.9\% relative improvement over the strongest baseline when using Mistral-Instruct and Llama-3-8B as the base models, respectively.

\paragraph{Organization.} Section~\ref{sec:related} presents related work on RLHF and learning in games. The problem formulation and preliminaries are provided in Section~\ref{sec:prelim}. Our algorithm and its theoretical guarantees are detailed in Section~\ref{sec:algo}. In Section~\ref{sec:discuss}, we compare our approach with other general preference alignment algorithms and explore its extension to the multi-turn setting. Experimental results are presented in Section~\ref{sec:exp}. Finally, we conclude the paper and discuss future directions in Section~\ref{sec:conclusion}.
\section{Related Work}\label{sec:related}
\paragraph{Reward-Based RLHF.} Since the first RLHF framework proposed by \citet{christiano2017deep}, RLHF has achieved tremendous success in aligning large language models (LLMs), powering models such as Instruct-GPT~\citep{ouyang2022training}, Llama 2~\citep{touvron2023llama}, and Claude~\citep{bai2022training}. The RLHF pipeline typically involves training a reward model followed by applying policy gradient methods such as PPO~\citep{schulman2017proximal} to optimize a KL-regularized objective~\citep{korbak2022rl,li2023remax}. Nevertheless, the use of PPO in RLHF introduces challenges, including instability during training~\citep{choshen2019weaknesses} and high computational costs~\citep{yuan2023rrhf}. To address these limitations, \citet{rafailov2024direct} proposed the DPO algorithm, which directly optimizes preferences by minimizing a loss objective on offline datasets. Additionally, other direct preference learning algorithms have been developed, including offline methods~\citep{ethayarajh2024kto} and online (iterative) methods~\citep{xie2024exploratory,xiong2024iterative,yuan2024self}. However, all these algorithms are reward-based and rely on the Bradley-Terry (BT) model assumption. In this paper, we remove the BT model assumption and consider general preference alignment.

\paragraph{RLHF with General Preferences.}\citet{azar2024general} is the first to consider the general preference without BT model assumption. They propose the offline IPO algorithm to learn the optimal policy when the comparator policy is fixed.~\citet{munos2023nash} formulate the alignment problem as a two-player zero-sum game and propose the iterative Nash-MD algorithm to find the Nash policy of the game. Subsequently, there has been a line of work~\citep{ye2024theoretical,calandriello2024human,rosset2024direct,wu2024self} developing online algorithms for learning the Nash policy. The closest work related to ours is~\citet{zhang2024iterative}, which also employs a no-regret learning algorithm for self-play. However, our algorithm incorporates an optimistic predictor into the policy update, achieving improved theoretical guarantees and better empirical performance. A detailed comparison between our algorithm and other general preference alignment algorithms is provided in Section~\ref{sec:discuss}.

\paragraph{Learning in Games.} Online learning and self-play algorithms are widely used in approximating the equilibrium of games, including normal-form games~\citep{freund1999adaptive,daskalakis2011near,mai2018cycles,roy2019online,chen2020hedging,wei2020linear,daskalakis2021near}, extensive-form games~\citep{zinkevich2007regret,kroer2020faster,kozuno2021model,lee2021last,bai2022near} and Markov games~\citep{wei2017online,jin2021v,liu2021sharp,mao2023provably}. Our work is inspired by the faster convergence properties of optimistic online mirror descent in equilibrium learning~\citep{rakhlin2013optimization,syrgkanis2015fast}.

\section{Preliminary}\label{sec:prelim}
\paragraph{Problem Setup.} We study the contextual formulation which is extensively used in previous RLHF literature~\citep{rafailov2024direct,xiong2024iterative}. The prompt $x \in \Xcal$ is sampled from an unknown prompt distribution $d_1$. $\Ycal$ is the response space and an LLM is characterized by a policy $\pi:\Xcal \rightarrow \Delta(\Ycal)$ which outputs the response probability given the context. For any policy $\pi$, we use $\E_{\pi}$ to denote the expectations under $\pi$. 

\paragraph{General Preferences.} In this work, we drop the BT model assumption~\citep{bradley1952rank} and focus on directly aligning LLMs with general preferences. To this end, we define a general preference oracle as follows:
\begin{definition}[General Preference Oracle] \label{def:general_oracle} There exists a preference oracle $\P: \Xcal \times \Ycal \to \Ycal \rightarrow [0,1]$, which can be queried to obtain the binary preference signal:
$$
z \sim \mathrm{Ber}\big(\P(y^1 \succ y^2 \mid x)),
$$
where $z=1$ indicates $y^1$ is preferred to $y^2$, and $z = 0$ indicates the opposite.
\end{definition}
Unlike the BT model assumption, which assumes the existence of a reward function $R^*$ for each $x$ and $y$, the general preference oracle always compares $y^1$ to another $y^2$. This setup aligns with practical scenarios, where it is often easier for users to compare two responses than to assign an absolute score to a single response. Since the preference signal always involves two responses, potentially come from two different policies, we formulate the LLM alignment problem as a two-player zero-sum game. The objective of this game is the expected win rate between the two players:
\begin{align*}
J(\pi_1,\pi_2):=\E_{x \sim d_1}\E_{y^1 \sim \pi_1,y^2 \sim \pi_2}\bra{\mathbb{P}(y^1 \succ y^2 \mid x)}.
\end{align*}
Here $\pi_1$ is the policy of the max-player, aiming to maximize the objective, while $\pi_2$ is the policy of the min-player, aiming to minimize it.


\paragraph{Nash Policies and Duality Gap.} Our learning goal is to find the Nash equilibrium of the game, which is defined as:
\begin{align*}
\pi_1^*,\pi_2^*:=\argmax_{\pi_1} \argmin_{\pi_2} J(\pi_1,\pi_2).
\end{align*}
Due to the symmetric nature of the game, the Nash policies for both players are identical, i.e., $\pi_1^*=\pi_2^*=\pi^*$, and the game value is $J(\pi^*,\pi^*)=0.5$. Since Nash policies are the best responses to each other, for any policy $\pi$, we have $J(\pi^*,\pi) \ge 0.5$, indicating that the Nash policy will not lose to any other policy. To quantify how well a policy $\pi$ approximates $\pi^*$, we define the duality gap as:
\begin{align*}
\mathrm{DualGap}(\pi):=\max_{\pi_1} J(\pi_1,\pi)-\min_{\pi_2} J(\pi, \pi_2).
\end{align*}
The duality gap is non-negative and $\mathrm{DualGap}(\pi) = 0$ if and only if $\pi = \pi^*$. Hence, our goal is to find a policy that minimizes the duality gap. Once we achieve $\mathrm{DualGap}(\pi) \leq \epsilon$, we say that $\pi$ is an $\epsilon$-approximate Nash policy.


\section{Algorithm}\label{sec:algo}
In this section, we begin by briefly reviewing the self-play algorithm with online mirror descent (OMD) updates, which is used in previous general preference alignment algorithm~\citep{zhang2024iterative}. Next, we present our proposed algorithm, which leverages the faster convergence properties of optimistic OMD, inspired by advancements in game theory~\citep{rakhlin2013optimization,syrgkanis2015fast}. Through theoretical analysis, we show that our approach achieves an improved bound on the duality gap. Finally, we describe the implementation of our algorithm. Following \citet{azar2024general,zhang2024iterative}, we omit the context $x$ throughout the rest of the paper since each context is independent.

\subsection{Self-play Algorithm with OMD Update}\label{sec:inpo}
Self-play algorithms are widely used in approximating the Nash policy~\citep{bai2020near,liu2021sharp}. The key idea is to let the policy play against itself, enabling iterative self-improvement. The algorithm is performed in an online manner, with each iteration using online mirror descent (OMD) to update the policy. Specifically, at iteration $t$, we find the policy that maximizes the following objective:
\begin{align}\label{eq:single_update}
\pi_{t+1}=\argmax_{\pi} \inner{\pi,r_t}-\frac{1}{\eta}\KL(\pi \Vert \pi_t),
\end{align}
where $r_t(y)=\mathbb{P}(y \succ \pi_t)=\E_{y' \sim \pi_t}[\mathbb{P}(y \succ y')]$ is the expected win rate of response $y$ against the current policy $\pi_t$, and $\eta>0$ is the learning rate. This objective ensures that $\pi_{t+1}$ not only aims to maximize the win rate over $\pi_t$ but also remains close to $\pi_t$, as measured by the KL divergence term. The stability introduced by the KL regularization is critical for achieving a sublinear regret bound. Without this regularization, one can construct examples where the algorithm suffers from linear regret, which is undesirable~\citep{lattimore2020bandit}.


Similar to the analysis in \citet{zhang2024iterative}, we can show that the uniform mixture of $\pi_{1:T}$ achieves an $\Ocal(T^{-1/2})$ duality gap, as stated in the following theorem. The proof is deferred to Appendix~\ref{sec:proof_omd}.
\begin{theorem}\label{thm:omd_guarantee}
Let $D=\max_{\pi} \KL(\pi \Vert \pi_1)$ and $\bar \pi=\frac{1}{T}\sum_{t=1}^T \pi_t$. Self-play algorithm in Eq.~\eqref{eq:single_update} with $\eta=\sqrt{\frac{D}{T}}$ satisfies:
\begin{align*}
\mathrm{DualGap}(\bar \pi) \le \frac{4\sqrt{D}}{\sqrt{T}}.
\end{align*}
\end{theorem}
\citet{zhang2024iterative} also demonstrate that self-play with OMD achieves last-iterate convergence. This result is attributed to the strong convexity induced by the KL regularization terms in their game objectives. However, since our objective does not include these KL terms, the last-iterate convergence may not hold in our game formulation.

\subsection{Optimistic Nash Policy Optimization}
While self-play with OMD update already achieves an $\Ocal(\sqrt{T})$ regret bound, which is near-optimal in many online learning scenarios, there is still room for improvement by better leveraging the self-play structure. Recent advancements in learning in games~\citep{rakhlin2013optimization, syrgkanis2015fast} demonstrate that a faster convergence rate of $\Ocal(T^{-1})$ can be achieved when both players adopt optimistic OMD update. In this subsection, we introduce how to integrate optimistic OMD into the self-play algorithm, resulting in an algorithm called Optimistic Nash Policy Optimization (ONPO).

The key idea of optimistic OMD is to incorporate a reward or loss predictor at each iteration. Recall that in OMD update, we use the expected win rate over the current policy $\pi_t$ as the reward vector $r_t$ to compute $\pi_{t+1}$. While in optimistic OMD, the learner utilizes a reward predictor $m_t$ and adopts a two-step update strategy:
\begin{align*}
\pi_t&=\argmax_{\pi} \inner{\pi,m_t}-\frac{1}{\eta}\KL(\pi \Vert \pi'_t)  \\
\pi'_{t+1}&=\argmax_{\pi} \inner{\pi,r_t}-\frac{1}{\eta}\KL(\pi \Vert \pi'_t).
\end{align*}
Here $\pi_t$ aims to maximize the reward predictor $m_t$ and the auxiliary policy $\pi'_{t+1}$ is updated after observing the actual reward $r_t$. The word ``optimistic'' comes from that the learner believes that the predictor $m_t$ provides a good approximation of the true reward $r_t$.

Next, we describe how to apply optimistic OMD in our self-play algorithm. In both OMD and optimistic OMD, the KL regularization term is consistently used to ensure that the next policy remains close to the previous policies. This regularization provides stability, making it reasonable to assume that the change from $\pi_t$ to $\pi_{t+1}$ is small. Based on this observation, we directly use the reward information from the previous iteration as the predictor, i.e., let $m_t=r_{t-1}=\E_{y
' \sim \pi_{t-1}}\bra{\mathbb{P}(y \succ y')}$.

In the following theorem, we demonstrate that ONPO achieves an $\Ocal(1/T)$ duality gap, improving over the previous $\Ocal(1/\sqrt{T})$ result. 
\begin{theorem}\label{thm:onpo_regret}
Let $D=\max_{\pi} \KL(\pi \Vert \pi'_1)$ and $\bar \pi=\frac{1}{T}\sum_{t=1}^T \pi_t$, ONPO algorithm with $\eta = \min\{\frac{1}{2},\sqrt{D}\}$ satisfies:
\begin{align*}
\mathrm{DualGap}(\bar \pi) \le \frac{4\sqrt{D}}{T}.
\end{align*}
\end{theorem}
Here, $\pi'_1 = \pi_1$ is the initialization policy. Theoretically, $\pi'_1$ can be set as a uniform policy, in which case $D$ is bounded by $\log |\mathcal{Y}|$. In RLHF practice, $\pi'_1$ is typically a supervised fine-tuned policy.

The proof is provided in Appendix~\ref{sec:proof_onpo}. The key to achieving the $\Ocal(1/T)$ rate lies in the regret bounded by variation in utilities (RVU) property of optimistic OMD. Specifically, the stability terms $\|r_t - r_{t-1}\|_{\infty}^2$ are canceled out by the negative term $-\|\pi_t - \pi_{t-1}\|_1^2$, which arises from the self-play mechanism where $r_t$ represents the win rate over $\pi_t$.  

Notably, the duality gap bound in \citet{zhang2024iterative} also depends on the maximum log density ratio between $\pi_t$ and a reference policy $\pi_{\mathrm{ref}}$, due to the KL-regularized game formulation. When optimistic OMD is applied in such a regularized game, the stability terms transform into  
$$
\max_y \left|\mathbb{P}(y \succ \pi_t) - \mathbb{P}(y \succ \pi_{t-1}) + \log \frac{\pi_t(y)}{\pi_{t-1}(y)}\right|,
$$ 
which cannot be canceled by the negative terms. However, the motivation behind regularizing the game is to keep the learner's policy close to the reference policy $\pi_{\mathrm{ref}}$, which aligns with the stability introduced in our update rule. Therefore, explicit regularization in our game objective is not necessary.

\subsection{Implementation of ONPO}
In this subsection, we describe the implementation of ONPO with query access to the preference oracle $\mathbb{P}$. The primary challenge in implementing ONPO lies in computing $r_t(y)$, which involves taking an expectation over the entire policy $\pi_t$. Fortunately, this challenge can be addressed by avoiding the direct estimation of $r_t(y)$ and instead relying on binary preference feedback between responses.

To achieves this, our goal is to design a loss function that does not involve $\mathbb{P}(y \succ \pi_t)$ for policy optimization. We focus on obtaining the loss objective for $\pi_t$ here and the derivation for $\pi'_t$ is similar. The key observation is that, $\pi_t$ has a closed-form solution which satisfies $\forall y,y' \in \Ycal$,
$$
\log \frac{\pi_t(y)}{\pi_t(y')}-\log \frac{\pi'_t(y)}{\pi'_t(y')}=\eta \pa{\mathbb{P}(y \succ \pi_{t-1}}-\mathbb{P}(y' \succ \pi_{t-1}).
$$
Therefore, similar to the techniques used in~\citet{azar2024general,zhang2024iterative}, solving $\pi_t$ is equivalent to finding the minimizer of the following loss function:
{\small
\begin{align*}
\E_{y,y' \sim \pi_{t-1}}\bra{\pa{g_t(\pi,y,y')-\eta\pa{\mathbb{P}(y \succ \pi_{t-1})-\mathbb{P}(y' \succ \pi_{t-1})}}^2}.
\end{align*}}
\noindent where $g_t(\pi,y,y')=\log \frac{\pi(y)}{\pi(y')}-\log \frac{\pi'_t(y)}{\pi'_t(y')}$. Since the inside win rate term is with respect to $\pi_{t-1}$ and we also have an expectation over $\pi_{t-1}$ outside, the loss function can be further written as
\begin{align*}
\E_{y,y' \sim \pi_{t-1},y_w,y_l \sim \lambda_p(y,y')}\bra{\pa{g_t(\pi,y_w,y_l)-\frac{\eta}{2}}^2},
\end{align*}
where $\lambda_p$ is the preference distribution~\citep{calandriello2024human}:
\begin{align*}
\lambda_p(y,y')=\begin{cases} (y,y') \quad \textrm{with probability $\P(y \succ y')$} \\
(y',y) \quad \textrm{with probability $1- \P(y \succ y')$.}
\end{cases}
\end{align*}
To calculate the loss function, we only need the access to sample from the current policy, which is standard and easy to implement in practice. Putting everything together, the implementation of ONPO is summarized in Algorithm~\ref{alg:onpo_prac}.

In the beginning, we initialize $\pi'_1$ and $\pi_1$ with the supervised fine-tuned policy $\pi_{\textrm{SFT}}$. At each iteration $t$, we sample responses from the current policy $\pi_t$ and use the preference feedback from the oracle $\mathbb{P}$ to construct the dataset $D_t$. Then we can directly minimize the corresponding loss functions on $D_t$ to find $\pi'_{t+1}$ and $\pi_{t+1}$ respectively. We use the last iteration policy $\pi_T$ as the output policy, which is consistent with online RLHF practice~\citep{dong2024rlhf,wu2024self,zhang2024iterative}.

\begin{algorithm}[t]
\caption{Implementation of ONPO
\label{alg:onpo_prac}
}
\begin{algorithmic}[1]
\STATE {\bfseries Input:} Number of iterations $T$, learning rate $\eta$, preference oracle $\mathbb{P}$, supervised fine-tuned policy $\pi_{\textrm{SFT}}$.
\STATE Initialize $\pi'_1 \leftarrow \pi_{\textrm{SFT}}$, $\pi_1 \leftarrow \pi_{\textrm{SFT}}$.
\FOR{iteration $t = 1,2,\dotsc,T-1$}
\STATE Sample response pairs from the current policy $\pi_t$: $\{y_1^{(i)},y_2^{(i)}\}_{i=1}^n \sim \pi_t$.
\STATE Construct preference dataset $D_t=\{y_w^{(i)},y_l^{(i)}\}_{i=1}^n$ with feedback from the oracle $\mathbb{P}$.
\STATE Calculate $\pi'_{t+1}$ as:
\begin{align*}
\pi'_{t+1}=\argmin_{\pi} \E_{y_w,y_l \sim D_t}\bra{\pa{g_{t}(\pi,y_w,y_l)-\frac{\eta}{2}}^2}.
\end{align*}
\STATE Calculate $\pi_{t+1}$ as:
\begin{align*}
\pi_{t+1}=\argmin_{\pi} \E_{y_w,y_l \sim D_t}\bra{\pa{g_{t+1}(\pi,y_w,y_l)-\frac{\eta}{2}}^2}.
\end{align*}
\ENDFOR
\STATE Output $\pi_T$.
\end{algorithmic}
\end{algorithm}

\section{Discussion}\label{sec:discuss}
In this section, we first discuss the differences between ONPO and other general preference alignment methods. Then we introduce how to extend ONPO to the multi-turn setting.
\subsection{Comparison between ONPO and Other General Preference Alignment Methods}
\paragraph{IPO.}\citet{azar2024general} is the first to address general preference alignment in LLMs. The optimization objective of IPO is:
\begin{align*}
\max_{\pi} \E_{y \sim \pi, y' \sim \mu}\bra{\mathbb{P}(y \succ y')}-\tau \KL(\pi \Vert \pi_{\textrm{ref}}),
\end{align*}
where $\mu$ is a fixed policy. From a game-theoretic perspective, the goal of IPO is to find the best response to $\mu$. However, this approach only ensures that the learned policy outperforms $\mu$, which leaves the possibility that another policy could outperform the learned policy. In contrast, our approach focuses on learning the Nash policy in a two-player game. This provides stronger theoretical guarantees, as the Nash policy will not lose to any other policy.

\paragraph{Nash-MD.}\citet{munos2023nash} is the first to formulate the alignment problem as a two-player zero-sum game. Their game objective includes KL regularization terms, which ensure that the player's policy remains close to the reference policy $\pi_{\textrm{ref}}$. The KL terms are weighted by a parameter $\tau$. They proposed an iterative algorithm, Nash-MD, to learn the Nash policy of the game. At each iteration $t$, the policy is updated as:
\begin{align*}
\pi_{t+1}=\argmax_{\pi} \mathbb{P}(\pi \succ \pi'_t)-\frac{1}{\eta_t} \KL(\pi,\pi'_t),
\end{align*}
where $\pi'_t$ is a geometric mixture policy of the current policy $\pi_t$ and the reference policy $\pi_{\textrm{ref}}$:
\begin{align*}
\pi'_t(y)=\frac{\pi_t(y)^{1-\eta_t \tau}\pi_{\textrm{ref}}(y)^{\eta_t \tau}}{\sum_{y'}{\pi_t(y')^{1-\eta_t \tau}\pi_{\textrm{ref}}(y')^{\eta_t \tau}}}.
\end{align*}
Nash-MD requires sampling from the mixture policy $\pi'_t$. However, the response space $\mathcal{Y}$ is often exponentially large, making the exact computation of $\pi'_t$ intractable. To address this, \citet{munos2023nash} propose sampling from an approximate policy. The theoretical guarantees of this approximation remain unclear. In contrast, our approach only requires sampling from the current policy $\pi_t$, which is straightforward to implement in practice.

\paragraph{Online IPO.}\citet{calandriello2024human} propose the online IPO population loss:
\begin{align*}
\mathop{\E}_{\substack{y, y' \sim {\mathrm{SG}[\pi]} \\ y_w, y_l \sim \lambda_p(y, y')}}\bra{\pa{\log{\frac{\pi(y_w)\pi_{\textrm{ref}}(y_l)}{\pi(y_l)\pi_{\textrm{ref}}(y_w)}}-\frac{1}{2 \tau}}^2},
\end{align*}
where $\mathrm{SG}$ is the stop-gradient operator, which prevents gradients from propagating through the data-generation process. Unlike the offline IPO approach, which always samples from a fixed policy $\mu$, online IPO leverages responses generated by the current policy $\pi$. 

Since the policy $\pi$ is updated throughout training, policy gradient methods are used to minimize the objective. However, as discussed earlier, policy gradient methods in RLHF have limitations, including being resource-intensive and unstable to train. In contrast, ONPO avoids these challenges by directly minimizing a loss function over a preference dataset, offering a more stable and efficient implementation.

\paragraph{DNO.} The theoretical version of DNO (Algorithm 1 in \citet{rosset2024direct}) relies on computing $r_t(y)=\E_{y' \sim \pi_t}\bra{\mathbb{P}(y \succ y')}$, which requires taking an expectation over the current policy $\pi_t$. This computation is challenging to implement in practice, so \citet{rosset2024direct} propose a practical version, DNO-Prct (Algorithm 2), where $\pi_{t+1}$ is updated as follows:
\begin{align*}
\argmax_{\pi} \E_{y_w,y_l \sim D_t} \log \bra{\sigma\pa{\eta \log\frac{\pi(y_w)\pi_t(y_l)}{\pi_t(y_w)\pi(y_l)}}}.
\end{align*}
When constructing the dataset $D_t$, only response pairs with large margins are selected. This selection is motivated by the fact that, to approximate DNO, the ideal condition is $\sigma(r_t(y_w) - r_t(y_l)) \approx 1$. However, this cannot be fully achieved since $r_t(y) \in [0, 1]$. Notably, the objective of DNO-Prct is identical to the DPO objective~\citep{rafailov2024direct}. Therefore, DNO-Prct can be viewed as an iterative version of DPO.

\paragraph{SPPO.}\citet{wu2024self} propose a self-play algorithm SPPO. The policy update in SPPO is:
{\small
\begin{align*}
\pi_{t+1}=\argmin_{\pi} \E_{y \sim \pi_t}\pa{\log \frac{\pi(y)}{\pi_t(y)} - \eta \pa{\widehat P(y \succ \pi_t) - \frac{1}{2}}}^2,
\end{align*}}
where $\wh{P}$ is a heuristic approximation of $\mathbb{P}(y \succ \pi_t)$. However, obtaining an accurate estimation of $\mathbb{P}(y \succ \pi_t)$ is challenging in practice. For example, Hoeffding’s inequality suggests that more than 100 queries are needed to ensure $\left|\mathbb{P}(y \succ \pi_t) - \widehat{P}(y \succ \pi_t)\right| \leq 0.1$. This requirement results in high annotation and computation costs, as 100 oracle queries are needed for a single response $y$. In contrast, ONPO bypasses the need to estimate $\mathbb{P}(y \succ \pi_t)$ and instead relies on binary preference signals between two responses.

\paragraph{INPO.}\citet{zhang2024iterative} propose a self-play algorithm, INPO, which employs OMD to iteratively update the policy, as described in Section~\ref{sec:inpo}. Leveraging the faster convergence properties of optimistic OMD, ONPO achieves an improved duality gap bound of $\Ocal(T^{-1})$, compared to the $\Ocal(T^{-1/2})$ bound of INPO.

\subsection{Extension to the Multi-Turn Setting}  
In this subsection, we describe how ONPO can be extended to the multi-turn setting, which is formulated as a contextual Markov decision process (CMDP)~\citep{shani2024multi}. The interaction between the LLM and the environment unfolds as follows: the LLM starts at a fixed initial state $s_1 \in \Scal$ and takes an action $y_1 \sim \pi(\cdot \mid s_1)$. The environment then transitions to the next state $s_2 \sim P(\cdot \mid s_1, y_1)$ according to the transition dynamics $P$, and the LLM subsequently takes action $y_2 \sim \pi(\cdot \mid s_2)$. This process repeats for $H$ steps, ultimately reaching the final state $s_{H+1}$. At the end of the interaction, the preference oracle compares two final states and provides a preference signal:  
$
z \sim \mathrm{Ber}\big(\P(s_{H+1}^1 \succ s_{H+1}^2 )\big).
$
This CMDP formulation effectively captures various LLM applications, including chatbot interactions and token-level MDPs~\citep{rafailov2024r}.

In the multi-turn setting, the challenge is that preferences are only provided for the final states, and there is no direct feedback for intermediate states. To address this, we use Q-value functions, which capture the long-term expected outcomes, in the optimization objective. For each state $s_h$, the update rule for $\pi_{t+1}(\cdot \mid s_h)$ is:
\begin{align*}
\argmax_{\pi} \inner{\pi,Q^{\pi_t,\pi_t}(s_h,\cdot)} - \frac{1}{\eta}\KL(\pi(\cdot \mid s_h) \Vert \pi_t(\cdot \mid s_h)),
\end{align*}
where $Q^{\pi_t,\pi_t}(s_h,y_h)=\E_{\pi_t}\bra{\Pcal(s_{H+1} \succ \pi_t) \mid s_h,y_h}$ and  $\Pcal(s \succ \pi_t)$ represents $\E_{\pi_t}\bra{\mathbb{P}(s \succ s_{H+1})}$. Here $\inner{\pi,Q^{\pi_t,\pi_t}(s_h,\cdot)}$ measures the probability of $\pi$ outperforming $\pi_t$ at state $s_h$. The update rule for $\pi'_{t+1}$ is similar, except that the KL divergence is computed between $\pi$ and $\pi'_t$.

The primary challenge in implementing ONPO in the multi-turn setting lies in the efficient estimation of $Q^{\pi_t,\pi_t}$.~\citet{shani2024multi} propose to use an actor-critic framework that employs policy-gradient methods such as PPO~\citep{schulman2017proximal} for policy optimization. However, policy-gradient methods are known to exhibit high variance and sensitivity to implementation details, leading to increased computational costs. In this paper, we focus on implementing ONPO in the single-turn setting and leave the implementation under the multi-turn setting for future work.

\section{Experiments}\label{sec:exp}
\begin{table*}[ht]
    \centering 
    \caption{Results on three benchmarks. ``ONPO+Mistral-It" refers to tuning the Mistral-Instruct model with ONPO, while ``ONPO+Llama-3-SFT" refers to tuning the Llama-3-SFT model with ONPO. Results where the baseline outperforms ONPO are underlined.}\label{tab:res_main}
    \vspace{5pt}
    \begin{tabular}{c|c|cccc}
    \toprule
    \textbf{Model}  &\textbf{Size} & \textbf{AlpacaEval 2.0} & \textbf{Arena-Hard} & \textbf{MT-Bench}  \\ \midrule
    Iterative DPO + Mistral-It & 7B& 32.0 & 22.2  & 7.35 \\
    SPPO + Mistral-It & 7B & 33.1 & 24.5 & 7.51 \\
    INPO + Mistral-It & 7B & 35.3 & 25.3 & 7.46 \\
    ONPO + Mistral-It & 7B & \textbf{42.8} & \textbf{29.7} & \textbf{7.68} \\
    \midrule
    \midrule
    Iterative DPO + Llama-3-SFT & 8B& 28.3 & 31.9  & 8.34 \\
    SPPO + Llama-3-SFT & 8B& 38.5 & 32.9 & 8.23 \\
    INPO + Llama-3-SFT & 8B & 44.2 & \underline{37.0} & 8.28 \\
    ONPO + Llama-3-SFT & 8B & \textbf{48.6} & \textbf{36.4} & \textbf{8.40} \\
    \midrule
    \midrule
    Llama-3-8B-it & 8B & 24.8 & 21.2 & 7.97\\
    Tulu-2-DPO-70B & 70B & 21.2 & 15.0 & 7.89 \\
    Llama-3-70B-it & 70B & 34.4  &41.1 & 8.95\\
    Mixtral-8x22B-it & 141B & 30.9 & 36.4 & 8.66  \\
    \midrule
    \midrule
    GPT-3.5-turbo-0613 & - & 22.7  & 24.8 & 8.39 \\
    GPT-4-0613 & - & 30.2 & 37.9 & 9.18 \\
    Claude-3-Opus & - & 40.5 & 60.4 & 9.00  \\
    GPT-4 Turbo (04/09) & - & 55.0 & 82.6 & -\\
    \bottomrule
    \end{tabular}
    \end{table*}
\subsection{Main Results} \paragraph{Experiment Setup.} We implement ONPO following the online RLHF workflow described in \citet{dong2024rlhf}. Two base models are used as the initial policy $\pi_1$: Llama-3-SFT\footnote{\url{https://huggingface.co/RLHFlow/LLaMA3-SFT}}, based on Llama-3-8B~\citep{dubey2024llama}, and Mistral-Instruct-v0.3\footnote{\url{https://huggingface.co/mistralai/Mistral-7B-Instruct-v0.3}}, an instruct fine-tuned version of the Mistral-7B-v0.3. For the general preference oracle, we use a pairwise preference model\footnote{\url{https://huggingface.co/RLHFlow/pair-preference-model-LLaMA3-8B}}, which demonstrates better performance compared to the BT reward model~\citep{zhang2024iterative}. Training details for the preference model are available in \citet{dong2024rlhf}.

At each iteration, the current policy generates $K = 8$ responses using a set of prompts\footnote{\url{https://huggingface.co/datasets/RLHFlow/prompt-collection-v0.1}}. To select $y_w$ (winner) and $y_l$ (loser), we follow the tournament approach in \citet{zhang2024iterative}, where the eight responses are compared pairwise to identify the winning and losing responses.

Since online or iterative alignment methods have been shown to outperform offline counterparts, we focus on comparing ONPO with other online methods for a fair evaluation. These include iterative DPO~\citep{dong2024rlhf}, SPPO~\citep{wu2024self} and INPO~\citep{zhang2024iterative}, where the latter two are general preference alignment approaches.

We evaluate the models on three representative benchmarks: AlpacaEval 2.0~\citep{li2023alpacaeval}, Arena-Hard~\citep{li2024live} and MT-Bench~\citep{zheng2024judging}. AlpacaEval 2.0 has 805 instructions from five datasets, including self-instruct test set~\citep{wang2022self}, Open Assistant test set, Anthropic's helpful test set~\citep{bai2022training}, Vicuna test set~\citep{zheng2024judging} and Koala test set~\citep{koala_blogpost_2023}. Arena-Hard includes 500 challenging user queries from Chatbot Arena. Both AlpacaEval 2.0 and Arena-Hard compare model-generated answers against reference answers from a baseline model, using GPT-4 Preview-1106 as the judge model. We report the win rate for Arena-Hard and the length-controlled (LC) win rate~\citep{dubois2024length} for AlpacaEval 2.0. MT-Bench consists of 80 multi-turn questions, where responses are rated by GPT-4 on a 1-10 scale, with the average rating reported.

\paragraph{Results.} The model performance is summarized in Table~\ref{tab:res_main}. Our results show that ONPO consistently outperforms or achieves comparable performance to the baselines across both base models. Among the three benchmarks, the length-controlled (LC) win rate in AlpacaEval 2.0 exhibits the highest 0.98 Spearman correlation with Chatbot Arena rankings~\citep{dubois2024length}. In this benchmark, ONPO outperforms the strongest baseline by a clear margin—achieving a 9.9\% improvement on Llama-3-SFT and a 21.2\% improvement on Mistral-It. These results align with our theoretical findings, demonstrating that ONPO benefits from an improved bound on the duality gap. We also compare ONPO with other LLMs that have significantly larger parameters, such as Llama-3-70B-it, Mixtral-8x22B-it and GPT-4-Turbo. Remarkably, our ONPO even outperforms models with at least nine times more parameters.

\subsection{More Results on Academic Tasks}
\begin{table*}[ht]
    \centering
    \caption{Model performance on more academic benchmarks (AVG: average).}\label{tab:academic}
    \vspace{5pt}
    \begin{tabular}{c|c|c|c|c|c|c|c}
    \toprule
    \textbf{Model} & \textbf{GPQA} & \textbf{Hellaswag} & \textbf{MMLU-Pro} & \textbf{Winogrande} & \textbf{TruthfulQA} & \textbf{GSM8K} &\textbf{AVG} \\ \midrule
    Mistral-It & 30.1 & 83.5 & 30.4 & 74.2 & 59.7 & 49.5 & 54.6 \\ 
    Iterative DPO & 29.6 & 83.3 & 28.0 & 75.1 & 64.0 & 45.7 & 54.3 \\ 
    SPPO & 28.7 & 83.5 & 28.1 & 73.9 & 66.4 & 49.9 & 55.1 \\
    INPO & 28.8 & 82.9 & 28.9 & 74.9 & 64.7 & 46.3 & 54.4  \\
    ONPO & 30.4 & 83.7 & 29.9 & 75.1 & 65.5 & 47.8 & \textbf{55.4}  \\
    \bottomrule
    \end{tabular}
\end{table*}
In this subsection, we evaluate the model’s reasoning and calibration abilities across six academic benchmarks: GPQA~\citep{rein2023gpqa} for graduate-level science question answering, MMLU-Pro~\citep{wang2024mmlu} for multitask language understanding, Hellaswag~\citep{zellers2019hellaswag} for commonsense inference, Winogrande~\citep{sakaguchi2021winogrande} for difficult commonsense reasoning, TruthfulQA~\citep{lin2021truthfulqa} to assess the model’s tendency to reproduce falsehoods, and GSM8K~\citep{cobbe2021training} for mathematical reasoning.

It is important to note that these benchmarks primarily evaluate a model’s intrinsic knowledge and capabilities, which are developed during the pre-training stage rather than the alignment stage. However, as observed in prior work~\citep{ouyang2022training,openai2023gpt}, alignment can sometimes have a negative impact on these abilities—a phenomenon known as the ``alignment tax". Therefore, our purpose in presenting these results is to verify that our alignment method preserves the model’s abilities rather than demonstrating performance improvements.

We show the results using Mistral-Instruct-v0.3 as the base model and compare ONPO with three baselines as well as the base model itself. The results in Table~\ref{tab:academic} show that ONPO achieves a slightly higher average performance than both the base model and the baselines, demonstrating that ONPO does not over-align the model and effectively preserves its intrinsic knowledge and abilities.

\subsection{Hyperparameter Sensitivity Analysis}
\begin{figure}[t]
\centering
\includegraphics[width=0.45\textwidth]{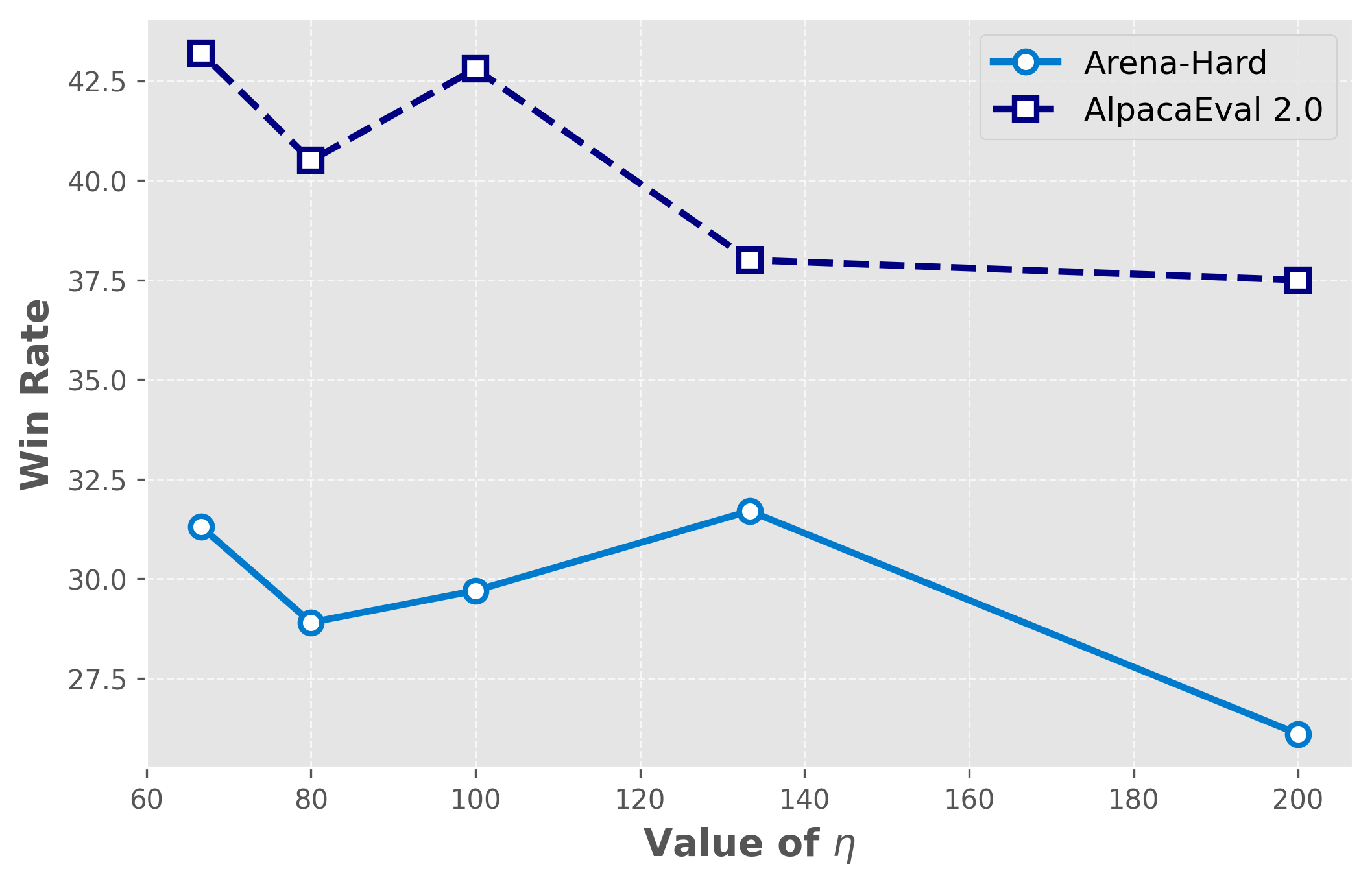}
\caption{
Performance of ONPO with different values of $\eta$ on Arena-Hard and AlpacaEval 2.0. ONPO consistently outperforms the best baseline, which achieves a win rate of 25.3 on Arena-Hard and 35.3 on AlpacaEval, respectively.}
\label{fig:res_eta}
\end{figure}
In this subsection, we analyze the sensitivity of ONPO to the hyperparameter $\eta$, which serves as the learning rate in the update rule. We conduct experiments using Mistral-Instruct-v0.3 as the base model and vary $\eta$ from $200/3$ to $200$. The results, presented in Figure~\ref{fig:res_eta}, indicate that ONPO consistently achieves strong performance across different values of $\eta$ and outperforms the baselines, demonstrating its robustness to hyperparamter variations.

\section{Conclusion and Future Work}\label{sec:conclusion}
We propose Optimistic Nash Policy Optimization (ONPO), a novel approach for aligning LLMs with general preferences via self-play. By integrating optimistic online mirror descent, ONPO achieves an improved duality gap bound for approximating the Nash policy of the game. Our experimental results demonstrate that ONPO consistently outperforms or matches state-of-the-art general preference alignment methods across multiple benchmarks. For future work, we aim to explore the implementation of ONPO under the multi-turn setting. In addition, we plan to design different strategies for actively selecting preference data to further enhance alignment performance.

\section*{Impact Statement}
This paper presents work whose goal is to advance the field of 
Machine Learning. There are many potential societal consequences 
of our work, none of which we feel must be specifically highlighted here.

\bibliography{ref}
\bibliographystyle{icml2025}

\newpage
\appendix
\onecolumn

\section{Proofs for Section~\ref{sec:algo}}\label{app:proof}

\subsection{Proof for Theorem~\ref{thm:omd_guarantee}}\label{sec:proof_omd}
\begin{proof}
According to the regret analysis of OMD~\citep{lattimore2020bandit}, for any policy $\pi$, we have
\begin{align*}
\sum_{t=1}^T \inner{\pi,r_t}-\sum_{t=1}^T \inner{\pi_t,r_t} &\le \frac{\KL(\pi \Vert \pi_1)}{\eta}+\eta \sum_{t=1}^T \|r_t\|^2_{\infty} \\
&\le 2 \sqrt{TD}.
\end{align*}
The rest proof follows from Theorem 3 in~\citet{zhang2024iterative}.
\end{proof}

\subsection{Proof for Theorem~\ref{thm:onpo_regret}}\label{sec:proof_onpo}
\begin{proof}
Let $\psi(\pi)=\sum_{y} \pi(y) \log \pi(y)$, the KL divergence between $\pi_1$ and $\pi_2$ can also be written as the Bregman divergence term:
\begin{align*}
\KL(\pi_1 \Vert \pi_2)=D_{\psi}(\pi_1,\pi_2)=\psi(\pi_1)-\psi(\pi_2)-\inner{\nabla \psi(\pi_2),\pi_1-\pi_2}.
\end{align*}
Since $\psi$ is strongly convex with respect to $L_1$ norm, we can apply regret analysis from~\citet{rakhlin2013optimization,syrgkanis2015fast} and obtain that for any $\pi'$
\begin{align*}
\sum_{t=1}^T \inner{\pi'-\pi_t,r_t} \le \frac{\KL(\pi' \Vert \pi'_1)}{\eta}+\eta \sum_{t=1}^T \|r_t-r_{t-1}\|^2_{\infty}-\frac{1}{4 \eta}\sum_{t=2}^T \|\pi_t-\pi_{t-1}\|_1^2.
\end{align*}
We observe that for any $t \ge 2$ and any $y$, 
$$
|r_t(y)-r_{t-1}(y)|=|\sum_{y'} \mathbb{P}(y \succ y')(\pi_t(y)-\pi_{t-1}(y))| \le \|\pi_t-\pi_{t-1}\|_1.$$
Once we have $\frac{1}{4 \eta} \ge \eta$, the terms $\eta \|r_t-r_{t-1}\|^2_{\infty}$ and $-\frac{1}{4 \eta}\|\pi_t-\pi_{t-1}\|^2_1$ cancel out and we get
\begin{align*}
\sum_{t=1}^T \inner{\pi'-\pi_t,r_t} \le 2 \sqrt{D}.
\end{align*}
Next, we decompose the duality gap as:
\begin{align*}
\mathrm{DualGap}(\bar \pi)=\underbrace{\max_{\pi_1} J(\pi_1,\bar \pi)-\frac{1}{2}}_{\textrm{Term A}}+\underbrace{\frac{1}{2}-\min_{\pi_2}J(\bar \pi, \pi_2)}_{\textrm{Term B}}.
\end{align*}
We show how to bound Term A and Term B is bounded similarly due to the symmetric nature of the game. Let $\pi'=\argmax_{\pi_1} J(\pi_1,\bar \pi)$, we have
\begin{align*}
J(\pi',\bar \pi)-\frac{1}{2}&=\frac{1}{T}\sum_{t=1}^T J(\pi',\pi_t)-J(\pi_t,\pi_t) \\
&=\frac{1}{T}\sum_{t=1}^T \inner{\pi'-\pi_t,r_t} \\
&\le\frac{2\sqrt{D}}{T}.
\end{align*}
The proof is finished by also having $\frac{1}{2}-\min_{\pi_2}J(\bar \pi, \pi_2) \le \frac{2\sqrt{D}}{T}$.
\end{proof}

\section{Additional Experiment Details}
For the implementation of ONPO, we follow the hyperparameters in \citet{dong2024rlhf}, including the cosine learning rate scheduler with a peak learning rate of $5 \times 10^{-7}$, a 0.03 warm-up ratio, and a global batch size of 128. We use a grid search for $1/\eta$ over $[0.1, 0.05, 0.02, 0.01, 0.005]$ and set $1/\eta = 0.01$. Llama-3-SFT is trained for 5 iterations, while Mistral-Instruct, having already undergone instruction fine-tuning, is thereby trained for 3 iterations.

\end{document}